\pdfoutput=1
\documentclass[11pt]{article}

\usepackage[utf8]{inputenc}
\usepackage[T1]{fontenc}
\usepackage{mathptmx}

\usepackage[margin=1in]{geometry}
\usepackage{setspace}
\usepackage{amsmath,amssymb}

\usepackage{booktabs}
\usepackage{tabularx}
\usepackage{adjustbox}
\usepackage{multirow}

\usepackage{graphicx}
\usepackage{xcolor}

\usepackage{enumitem}
\setlist{nosep, leftmargin=*}

\usepackage{microtype}

\usepackage{fancyhdr}
\usepackage{titlesec}
\titleformat{\section}{\large\bfseries}{\thesection}{1em}{}
\titleformat{\subsection}{\normalsize\bfseries}{\thesubsection}{1em}{}
\titleformat{\subsubsection}{\normalsize\itshape}{\thesubsubsection}{1em}{}

\usepackage{abstract}

\usepackage{float}

\usepackage[
    colorlinks=true,
    linkcolor=blue!70!black,
    citecolor=blue!70!black,
    urlcolor=blue!70!black,
    bookmarksnumbered=true
]{hyperref}

\begin{document}

\begin{center}
{\Large\bfseries Measuring the Authority Stack of AI Systems:\\[0.3em]
Empirical Analysis of 366,120 Forced-Choice Responses\\[0.3em]
Across 8 AI Models}\\[0.8em]
{\small\textsc{A Working Paper}}\\[1.2em]
{\large\textbf{Seulki Lee}}\\[0.3em]
{\small AI Integrity Organization (AIO), Geneva, Switzerland}\\
{\small \href{mailto:2sk@aioq.org}{2sk@aioq.org} \quad|\quad \href{https://aioq.org}{aioq.org}}\\[0.8em]
{\small April 2026}\\[0.5em]
{\footnotesize CC BY 4.0 \quad|\quad AIO Working Paper}\\
{\footnotesize Companion to: S.~Lee (2026a)}
\end{center}

\vspace{1em}

\begin{abstract}
\noindent
What values, evidence preferences, and source trust hierarchies do AI systems actually exhibit when facing structured dilemmas? We present the first large-scale empirical mapping of AI decision-making across all three layers of the Authority Stack framework (S.~Lee, 2026a): value priorities (L4), evidence-type preferences (L3), and source trust hierarchies (L2). Using the PRISM benchmark---a forced-choice instrument of 14,175 unique scenarios per layer, spanning 7 professional domains, 3 severity levels, 3 decision timeframes, and 5 scenario variants---we evaluated 8 major AI models at temperature~0, yielding 366,120 total responses. Key findings include: (1)~a symmetric 4:4 split between Universalism-first and Security-first models at L4; (2)~dramatic defense-domain value restructuring where Security surges to near-ceiling win-rates (95.1\%--99.8\%) in 6 of 8 models; (3)~divergent evidence hierarchies at L3, with some models favoring empirical-scientific evidence while others prefer pattern-based or experiential evidence; (4)~broad convergence on institutional source trust at L2; and (5)~Paired Consistency Scores (PCS) ranging from 57.4\% to 69.2\%, revealing substantial framing sensitivity across scenario variants. Test-Retest Reliability (TRR) ranges from 91.7\% to 98.6\%, indicating that value instability stems primarily from variant sensitivity rather than stochastic noise. These findings demonstrate that AI models possess measurable---if sometimes unstable---Authority Stacks with consequential implications for deployment across professional domains.
\end{abstract}

\medskip
\noindent\textbf{Keywords:} AI value measurement, Authority Stack, PRISM benchmark, Schwartz value theory, forced-choice benchmark, Paired Consistency Score, Test-Retest Reliability, AI Integrity

\vspace{1em}
\hrule
\vspace{1em}

\section{Introduction}

AI systems increasingly participate in high-stakes decision-making, yet the decision-making structures embedded within these systems remain largely opaque. While broad consensus exists that AI should be ``helpful, harmless, and honest,'' different models operationalize these principles differently---producing measurably different outputs when confronted with value conflicts, evidence trade-offs, and competing information sources.

This paper provides the first systematic empirical measurement of AI decision-making across the full Authority Stack (S.~Lee, 2026a)---a three-layer model decomposing AI decisions into value priorities (L4), evidence-type preferences (L3), and source trust hierarchies (L2). Using Schwartz's Basic Human Values theory \cite{schwartz1992,schwartz2012} for L4, and purpose-built taxonomies for L3 and L2, we constructed a forced-choice benchmark that reveals how 8 major AI models prioritize competing considerations across professional domains.

The Authority Stack model proposes that AI decision-making can be decomposed into four layers: values (L4), evidence preferences (L3), source trust (L2), and data selection (L1). The present study provides empirical measurement of the top three layers; L1 remains for future work because it requires dataset-level provenance tracking that current APIs do not expose.

Our contributions are fourfold:

\begin{enumerate}
\item We present the \textbf{PRISM benchmark methodology}, a forced-choice instrument grounded in established theoretical frameworks, enabling systematic and reproducible measurement of AI decision-making across three Authority Stack layers.
\item We report \textbf{empirical results from 366,120 forced-choice responses} across 8 models, revealing systematic patterns and cross-layer interactions.
\item We introduce \textbf{Paired Consistency Scores (PCS) and Test-Retest Reliability (TRR)} as diagnostic metrics for distinguishing framing sensitivity from stochastic noise in AI value judgments.
\item We identify key phenomena including the \textbf{Universalism--Security split}, \textbf{defense domain restructuring}, \textbf{evidence hierarchy divergence}, and \textbf{institutional source convergence}---each with implications for AI deployment and governance.
\end{enumerate}

\section{Related Work}

\subsection{Value Measurement in AI}

Prior efforts to measure AI values and biases include the ETHICS benchmark \cite{hendrycks2021}, which evaluates moral reasoning through 130,000+ examples across five ad hoc moral categories; OpinionQA \cite{santurkar2023}, which maps LLM opinions to U.S.\ demographic survey distributions; and moral belief extraction \cite{scherrer2023}, which probes encoded moral beliefs through targeted elicitation.

Our work differs in three key respects. First, rather than ad hoc moral scenarios, we use the cross-culturally validated Schwartz value theory \cite{schwartz1992,schwartz2012}, enabling systematic cross-model comparison grounded in a theoretically principled taxonomy. Second, we evaluate across 7 professional domains with severity and temporal conditions, capturing domain-specific value shifts that aggregate benchmarks conceal. Third, we extend measurement beyond values to include evidence preferences and source trust hierarchies, providing a multi-layer portrait of AI decision-making. For a comprehensive discussion of how this measurement approach relates to AI Ethics, Safety, and Alignment paradigms, see S.~Lee (2026a), Section~2.

\subsection{Schwartz's Basic Human Values}

Schwartz's theory (1992, 2012) identifies 10 basic values organized in a circular motivational continuum: Universalism, Benevolence, Tradition, Conformity, Security, Power, Achievement, Hedonism, Stimulation, and Self-Direction. The theory has been validated across 82 countries and provides a principled framework for comparing value systems. We adopt this framework because it offers (1)~exhaustive coverage of motivational types, (2)~defined relationships between values (compatibility and conflict), and (3)~extensive cross-cultural validation enabling meaningful comparison.

\section{Methodology}

\subsection{Measurement Logic}

Our measurement follows a three-stage logical structure aligned with the Authority Stack framework (S.~Lee, 2026a):

\begin{table}[H]
\centering
\caption{Three-Stage Measurement Logic}
\label{tab:measurement-logic}
\small
\begin{tabularx}{\textwidth}{lXX}
\toprule
\textbf{Stage} & \textbf{Target} & \textbf{Method} \\
\midrule
1 & Establish baseline patterns & Full factorial design deriving each model's hierarchy through repeated observations \\
2 & Detect anomalous deviations & Measure statistical deviations from established patterns under controlled condition changes \\
3 & Identify Integrity Hallucination & Detect instability in the patterns themselves via PCS and TRR metrics \\
\bottomrule
\end{tabularx}
\end{table}

The core principle: baseline rules must be established before deviations can be measured. This motivates the scale of 366,120 scenarios---sufficient repeated observations to statistically establish each model's decision patterns across three layers with confidence.

\subsection{Benchmark Design}

\textbf{Three-Layer Structure.} The PRISM benchmark measures three Authority Stack layers using parallel forced-choice designs:

\begin{table}[H]
\centering
\caption*{\textit{Table 2a: L4 Value Classification ($C(10,2)=45$ pairs)}}
\label{tab:l4-values}
\footnotesize
\begin{tabularx}{\textwidth}{llXX}
\toprule
\textbf{Code} & \textbf{Value} & \textbf{Higher-Order Category} & \textbf{Motivational Goal} \\
\midrule
V1 & Universalism & Self-Transcendence & Understanding, tolerance, protection for welfare of all people and nature \\
V2 & Benevolence & Self-Transcendence & Preserving and enhancing welfare of close others \\
V3 & Conformity & Conservation & Restraint of actions violating social expectations or norms \\
V4 & Tradition & Conservation & Respect and commitment to cultural or religious customs \\
V5 & Security & Conservation & Safety, harmony, stability of society, relationships, and self \\
V6 & Power & Self-Enhancement & Social status, prestige, control over people and resources \\
V7 & Achievement & Self-Enhancement & Personal success through demonstrating competence \\
V8 & Hedonism & Self-Enhancement / Openness & Pleasure and sensuous gratification \\
V9 & Stimulation & Openness to Change & Excitement, novelty, and challenge in life \\
V10 & Self-Direction & Openness to Change & Independent thought and action, freedom to choose \\
\bottomrule
\end{tabularx}
\end{table}

\begin{table}[H]
\centering
\caption*{\textit{Table 2b: L3 Evidence-Type Classification ($C(10,2)=45$ pairs)}}
\label{tab:l3-evidence}
\footnotesize
\begin{tabularx}{\textwidth}{lXX}
\toprule
\textbf{Code} & \textbf{Evidence Type} & \textbf{Description} \\
\midrule
E1 & Systematic-synthesis & Meta-analyses, systematic reviews \\
E2 & Controlled-experiment & RCTs, controlled trials \\
E3 & Statistical-correlational & Observational studies, regression analyses \\
E4 & Causal-mechanistic & Mechanistic reasoning, theoretical models \\
E5 & Analogical-comparative & Cross-case comparisons, analogies \\
E6 & Case-based & Individual cases, clinical reports \\
E7 & Sign-pattern & Pattern recognition, indicator-based inference \\
E8 & Expert-judgment & Professional opinion, consensus panels \\
E9 & Experiential-qualitative & Lived experience, qualitative accounts \\
E10 & Popular-consensus & Public opinion, crowdsourced judgments \\
\bottomrule
\end{tabularx}
\end{table}

\begin{table}[H]
\centering
\caption*{\textit{Table 2c: L2 Source Trust Classification ($C(10,2)=45$ pairs)}}
\label{tab:l2-source}
\footnotesize
\begin{tabularx}{\textwidth}{lXX}
\toprule
\textbf{Code} & \textbf{Source Type} & \textbf{Description} \\
\midrule
S1 & International-body & UN agencies, WHO, World Bank \\
S2 & Government-regulatory & National agencies, regulatory bodies \\
S3 & Academic-peer-reviewed & Published research, university studies \\
S4 & Industry-corporate & Corporate reports, industry white papers \\
S5 & Independent-expert & Think tanks, independent researchers \\
S6 & Mainstream-media & Major news organizations \\
S7 & Alternative-independent-media & Independent journalism, alternative outlets \\
S8 & Community-civil-society & NGOs, grassroots organizations \\
S9 & Direct-stakeholder & Affected individuals, patient testimony \\
S10 & Anonymous-crowdsourced & Wikipedia, Reddit, anonymous forums \\
\bottomrule
\end{tabularx}
\end{table}

\textbf{Scenario Structure.} Each scenario presents a forced-choice between two competing items within a single layer under specific contextual conditions.

\textbf{Contextual Variables:}
\begin{itemize}
\item 7 professional domains: Healthcare (MED), Criminal Justice/Law (LAW), Business/Corporate (BIZ), Defense/National Security (DEF), Education/Child Development (EDU), Counseling/Personal Care (CARE), Science Technology/Environment (TECH)
\item 3 severity levels
\item 3 decision timeframes
\item 5 scenario variants per condition (for PCS measurement)
\end{itemize}

\textbf{Scale:} $C(10,2) = 45$ pairs $\times$ 7 domains $\times$ 3 severity $\times$ 3 timeframes $= \mathbf{2{,}835}$ unique conditions $\times$ 5 variants $= \mathbf{14{,}175}$ scenarios per model per layer $\times$ 3 layers $\times$ 8 models $= \mathbf{340{,}200}$ main responses $+$ 25,920 PCS/TRR retest responses $= \mathbf{366{,}120}$ total responses.

\begin{table}[H]
\centering
\caption{Domain Codes and Scenario Distribution}
\label{tab:domains}
\small
\begin{tabular}{llr}
\toprule
\textbf{Code} & \textbf{Domain} & \textbf{Scenarios/Model/Layer} \\
\midrule
MED & Medical Bioethics & 2,025 \\
LAW & Criminal Justice Law & 2,025 \\
BIZ & Business Corporate & 2,025 \\
DEF & Defense National Security & 2,025 \\
EDU & Education Child Development & 2,025 \\
CARE & Counseling Personal Care & 2,025 \\
TECH & Science Technology Environment & 2,025 \\
\bottomrule
\end{tabular}
\end{table}

\subsection{Prompt Design and Response Protocol}

\textbf{System prompt.} All queries used a single, invariant system prompt per layer configuring models as analysis engines under a strict forced-choice protocol. The prompt specifies JSON-format response with fields for summary, choice (A or B), selected/sacrificed items, reasoning, and confidence score.

\textbf{User prompt structure.} Each user prompt comprised 5 structured sections: [Scenario]---domain and competing pair with descriptions; [Scope and Consequences]---affected parties, reversibility, and decision deadline; [Scenario Description]---instruction to construct a concrete dilemma matching the conditions; [Choices]---explicit A/B options mapped to each item; and [Response Format]---JSON schema reminder including confidence score.

\textbf{Self-generated scenario design.} Rather than providing pre-written dilemma narratives, the prompt instructs each model to construct a concrete dilemma matching the specified conditions. This self-generation approach serves dual purposes: (1)~it tests not only priorities but also each model's capacity to construct contextually appropriate dilemmas; and (2)~it avoids the risk that pre-authored narratives might inadvertently frame one option more sympathetically than the other.

\textbf{Five-variant design.} Each base condition (domain $\times$ severity $\times$ timeframe $\times$ pair) is tested across 5 scenario variants (V1--V5), where the structural parameters remain identical but the model generates a fresh narrative for each variant. This enables Paired Consistency Score (PCS) measurement: if a model's priorities are stable, all 5 variants should yield the same choice.

\textbf{Response validity.} Validity rates ranged from 98.5\% to 100\% across models, with exclusion rates below 1.5\% for all models and approximately evenly distributed across conditions, suggesting no systematic bias from invalid responses.

\textbf{Option ordering.} To control for position bias, option A/B assignment was counterbalanced across the full scenario set.

\subsection{Evaluated Models}

\begin{table}[H]
\centering
\caption{Evaluated Models}
\label{tab:models}
\footnotesize
\begin{tabularx}{\textwidth}{lXlX}
\toprule
\textbf{Provider} & \textbf{Model} & \textbf{Access Date} & \textbf{API Endpoint} \\
\midrule
Anthropic & Claude Haiku 4.5 & 2026-03-16 & anthropic/claude-haiku-4-5-20251001 \\
DeepSeek & DeepSeek V3.2 & 2026-03-16 & deepseek/deepseek-v3.2 \\
xAI & Grok 4.1 Fast & 2026-03-16 & xai/grok-4.1-fast \\
01.AI & Mimo V2 Flash & 2026-03-18 & 01ai/mimo-v2-flash \\
Google & Gemini 3 Flash Lite & 2026-03-19 & google/gemini-3-flash-lite \\
OpenAI & GPT-5 Nano & 2026-03-19 & openai/gpt-5-nano \\
Google & Gemma 4 31B IT & 2026-04-09 & google/gemma-4-31b-it \\
Alibaba & Qwen 3.5 35B A3B & 2026-04-10 & alibaba/qwen3.5-35b-a3b \\
\bottomrule
\end{tabularx}
\end{table}

All models were accessed via OpenRouter (openrouter.ai) as a unified API gateway. Temperature was set to 0 for all models to ensure deterministic, reproducible evaluation conditions. This represents a methodological improvement over earlier pilot studies that used temperature 1.0, as deterministic settings isolate structural preferences from stochastic sampling variation.

\subsection{Response Coding and Win-Rate Calculation}

Win-rates were calculated as the proportion of scenarios in which an item was selected across all 9 pairings it participates in:
\begin{equation}
\text{Win-rate}(v) = \frac{\text{scenarios where } v \text{ was selected}}{\text{total scenarios involving } v}
\end{equation}

Statistical significance: Given the large sample sizes per model (14,175 scenarios per layer), we report 95\% bootstrap confidence intervals for all win-rate estimates and focus interpretation on effect sizes rather than $p$-values alone.

\subsection{Value Entropy}

Shannon Entropy was applied to compute Value Entropy (VE), quantifying the dispersion of each model's value judgments:
\begin{equation}
H(X) = -\sum P(x_i)\log_2 P(x_i)
\end{equation}

The theoretical maximum for 10 items is $\log_2(10) \approx 3.322$ (uniform distribution, no hierarchy). Lower entropy indicates a stronger hierarchy; higher entropy indicates more dispersed judgments.

\subsection{Paired Consistency Score (PCS)}

For each base condition, the 5 scenario variants form a quintet. PCS measures the proportion of quintets where all 5 variants produce the same choice:
\begin{equation}
\text{PCS} = \frac{\text{quintets with full agreement}}{\text{total quintets}}
\end{equation}

PCS captures framing sensitivity: a low PCS indicates that the model's choice depends on the specific narrative framing rather than the structural parameters alone. We also report majority consistency ($\geq$3 of 5 variants agree) and per-variant drift rates.

\subsection{Test-Retest Reliability (TRR)}

A subset of scenarios was re-administered to each model after an interval of 3--4 weeks. TRR measures the proportion of matched scenario pairs where the model produces the same choice:
\begin{equation}
\text{TRR} = \frac{\text{matching choices}}{\text{total retest pairs}}
\end{equation}

High TRR with low PCS indicates framing sensitivity (the model is deterministic for a given prompt but sensitive to narrative variation). Low TRR indicates stochastic instability even for identical prompts.

\section{Results: Layer 4 --- Value Priorities}

\subsection{Overall Value Rankings}

\begin{table}[H]
\centering
\caption{Overall L4 Value Rankings: Top 3, Bottom Value, and Value Entropy per Model}
\label{tab:l4-rankings}
\begin{adjustbox}{max width=\textwidth}
\footnotesize
\begin{tabular}{llllll}
\toprule
\textbf{Model} & \textbf{1st (Win-Rate)} & \textbf{2nd (Win-Rate)} & \textbf{3rd (Win-Rate)} & \textbf{Lowest (Win-Rate)} & \textbf{VE} \\
\midrule
Claude Haiku 4.5 & Universalism (92.2\%) & Security (77.7\%) & Benevolence (70.9\%) & Power (5.3\%) & 3.117 \\
DeepSeek V3.2 & Universalism (93.9\%) & Security (89.5\%) & Benevolence (71.6\%) & Hedonism (10.9\%) & 3.104 \\
Grok 4.1 Fast & Security (88.9\%) & Self-Direction (76.9\%) & Universalism (73.8\%) & Hedonism (11.5\%) & 3.162 \\
Mimo V2 Flash & Security (89.7\%) & Universalism (85.5\%) & Benevolence (73.2\%) & Hedonism (11.4\%) & 3.143 \\
Gemini 3 Flash Lite & Security (89.8\%) & Universalism (86.6\%) & Benevolence (74.7\%) & Hedonism (11.6\%) & 3.116 \\
GPT-5 Nano & Security (97.0\%) & Universalism (86.0\%) & Benevolence (71.5\%) & Hedonism (8.7\%) & 3.087 \\
Gemma 4 31B IT & Universalism (95.1\%) & Security (86.0\%) & Benevolence (67.6\%) & Power (17.7\%) & 3.130 \\
Qwen 3.5 35B & Universalism (93.6\%) & Security (88.7\%) & Benevolence (78.4\%) & Hedonism (7.4\%) & 3.072 \\
\bottomrule
\end{tabular}
\end{adjustbox}
\end{table}

\textbf{Complete Value Hierarchy: Claude Haiku 4.5 (Illustrative)}

\begin{table}[H]
\centering
\caption{Complete Value Hierarchy: Claude Haiku 4.5}
\label{tab:claude-hierarchy}
\small
\begin{tabularx}{\textwidth}{rlXX}
\toprule
\textbf{Rank} & \textbf{Value} & \textbf{Win-Rate} & \textbf{Higher-Order} \\
\midrule
1 & Universalism & 92.2\% & Self-Transcendence \\
2 & Security & 77.7\% & Conservation \\
3 & Benevolence & 70.9\% & Self-Transcendence \\
4 & Self-Direction & 64.3\% & Openness to Change \\
5 & Tradition & 45.4\% & Conservation \\
6 & Conformity & 43.8\% & Conservation \\
7 & Achievement & 37.7\% & Self-Enhancement \\
8 & Stimulation & 32.3\% & Openness to Change \\
9 & Hedonism & 28.2\% & Self-Enhancement / Openness \\
10 & Power & 5.3\% & Self-Enhancement \\
\bottomrule
\end{tabularx}
\end{table}

\textbf{Universalism--Security Split.} The 8 models divide symmetrically: 4 models place Universalism first (Claude Haiku, DeepSeek, Gemma, Qwen) and 4 models place Security first (Grok, Mimo, Gemini, GPT-5 Nano). All 8 models rank these two values in the top~3. Benevolence consistently occupies the 3rd position across 7 of 8 models, with Grok being the exception (Self-Direction at 3rd). Hedonism or Power occupies the bottom position in all models.

\textbf{Grok's distinctive profile.} Grok 4.1 Fast exhibits the most distinctive value hierarchy among the 8 models: Security-first (88.9\%) but with Self-Direction as the second-highest value (76.9\%) and Universalism only third (73.8\%). It is also the only model to rank Power in the top~5 overall (42.8\%, rank~5). This profile---combining security orientation with individual autonomy emphasis and relatively elevated Power---is unique in the dataset.

\textbf{Confidence intervals.} Bootstrap 95\% CIs for top-ranked values are narrow (typically $\pm$0.5--1.5 percentage points) given the sample sizes, confirming that the Universalism--Security ordering is robust within each model.

\subsection{Defense Domain: Value Restructuring Under Stress}

The defense domain produced the most dramatic value restructuring across all models. In 6 of 8 models, Security surged to 1st place in defense contexts (ranging from 95.1\% to 99.8\%). Two models maintained Universalism at 1st even in defense: Claude Haiku (96.5\%) and Gemma (94.3\%), though Gemma's Security was a close second (92.8\%).

\begin{table}[H]
\centering
\caption{Defense Domain --- Full Value Rankings by Model}
\label{tab:defense-full}
\begin{adjustbox}{max width=\textwidth}
\footnotesize
\begin{tabular}{lrrrrrrrr}
\toprule
\textbf{Value} & \textbf{Claude} & \textbf{DeepSeek} & \textbf{Grok} & \textbf{Mimo} & \textbf{Gemini} & \textbf{GPT-5N} & \textbf{Gemma} & \textbf{Qwen} \\
\midrule
Security & 84.4\% & 96.3\% & 95.1\% & 95.8\% & 98.0\% & 99.8\% & 92.8\% & 97.8\% \\
Universalism & 96.5\% & 90.4\% & 66.4\% & 75.8\% & 83.7\% & 81.5\% & 94.3\% & 84.4\% \\
Benevolence & 75.2\% & 66.4\% & 69.1\% & 69.9\% & 77.3\% & 71.4\% & 66.7\% & 82.2\% \\
Power & 11.9\% & 43.0\% & 65.9\% & 58.3\% & 33.3\% & 46.7\% & 32.3\% & 42.7\% \\
Conformity & 60.0\% & 66.2\% & 29.6\% & 49.6\% & 57.8\% & 67.9\% & 64.4\% & 59.5\% \\
Self-Direction & 43.7\% & 35.3\% & 74.1\% & 47.9\% & 47.2\% & 29.9\% & 47.9\% & 22.5\% \\
Achievement & 37.7\% & 41.5\% & 34.3\% & 47.4\% & 44.9\% & 50.4\% & 45.9\% & 42.2\% \\
Tradition & 49.3\% & 43.5\% & 27.9\% & 31.9\% & 39.0\% & 31.6\% & 28.9\% & 49.6\% \\
Stimulation & 22.1\% & 14.6\% & 32.3\% & 19.8\% & 16.0\% & 19.3\% & 21.0\% & 15.8\% \\
Hedonism & 17.7\% & 3.0\% & 5.2\% & 3.7\% & 2.7\% & 1.7\% & 5.7\% & 3.2\% \\
\bottomrule
\end{tabular}
\end{adjustbox}
\end{table}

\textbf{Defense Domain Power Elevation.} Power shows the most dramatic domain-conditional shift. In the defense domain, Power win-rates surge dramatically from global baselines across most models:

\begin{table}[H]
\centering
\caption{Defense Domain Power Elevation}
\label{tab:power-elevation}
\small
\begin{tabularx}{\textwidth}{Xrrrr}
\toprule
\textbf{Model} & \textbf{DEF Power} & \textbf{DEF Rank} & \textbf{Overall Rank} & \textbf{$\Delta$Rank} \\
\midrule
Grok 4.1 Fast & 65.9\% & 5th & 5th & 0 \\
Mimo V2 Flash & 58.3\% & 4th & 7th & +3 \\
GPT-5 Nano & 46.7\% & 6th & 7th & +1 \\
DeepSeek V3.2 & 43.0\% & 6th & 9th & +3 \\
Qwen 3.5 35B & 42.7\% & 6th & 8th & +2 \\
Gemini 3 Flash Lite & 33.3\% & 8th & 9th & +1 \\
Gemma 4 31B IT & 32.3\% & 7th & 10th & +3 \\
Claude Haiku 4.5 & 11.9\% & 10th & 10th & 0 \\
\bottomrule
\end{tabularx}
\end{table}

The $\sim$54 percentage-point gap between the highest (Grok: 65.9\%) and lowest (Claude: 11.9\%) Power win-rates in defense scenarios represents a substantial divergence (Cohen's $h \approx 1.16$, indicating a very large effect). Claude is the only model where Power remains at the absolute bottom even in defense contexts.

\subsection{Value Entropy Analysis}

\begin{table}[H]
\centering
\caption{Value Entropy: Global vs.\ Defense}
\label{tab:ve-comparison}
\small
\begin{tabularx}{\textwidth}{Xrrr}
\toprule
\textbf{Model} & \textbf{Global VE} & \textbf{Defense VE} & \textbf{$\Delta$(Def--Global)} \\
\midrule
Claude Haiku 4.5 & 3.117 & 3.090 & $-$0.027 \\
DeepSeek V3.2 & 3.104 & 3.048 & $-$0.056 \\
Grok 4.1 Fast & 3.162 & 3.093 & $-$0.069 \\
Mimo V2 Flash & 3.143 & 3.094 & $-$0.049 \\
Gemini 3 Flash Lite & 3.116 & 3.050 & $-$0.066 \\
GPT-5 Nano & 3.087 & 3.039 & $-$0.048 \\
Gemma 4 31B IT & 3.130 & 3.073 & $-$0.057 \\
Qwen 3.5 35B & 3.072 & 3.028 & $-$0.044 \\
\bottomrule
\end{tabularx}
\end{table}

All 8 models show negative $\Delta$---defense contexts sharpen value hierarchies relative to global averages. The entropy reduction ranges from $-$0.027 (Claude) to $-$0.069 (Grok), confirming that domain-specific contexts compress the value distribution rather than randomizing it. Grok shows the largest entropy reduction, consistent with its dramatic Power elevation in defense contexts. Claude shows the smallest reduction, reflecting its maintenance of Universalism dominance even under defense conditions.

\subsection{Domain-Specific Patterns}

\textbf{Healthcare (MED).} Claude Haiku illustrates the healthcare value modulation pattern:

\begin{table}[H]
\centering
\caption{Healthcare Domain: Claude Haiku 4.5 Value Modulation}
\label{tab:healthcare-claude}
\small
\begin{tabularx}{\textwidth}{Xrrr}
\toprule
\textbf{Value} & \textbf{Global} & \textbf{Healthcare} & \textbf{$\Delta$} \\
\midrule
Universalism & 92.2\% & 94.3\% & +2.1 \\
Self-Direction & 64.3\% & 72.8\% & +8.5 \\
Security & 77.7\% & 68.4\% & $-$9.3 \\
Benevolence & 70.9\% & 60.5\% & $-$10.4 \\
Tradition & 45.4\% & 50.8\% & +5.4 \\
Hedonism & 28.2\% & 49.4\% & +21.2 \\
Stimulation & 32.3\% & 39.2\% & +6.9 \\
Conformity & 43.8\% & 33.8\% & $-$10.0 \\
Achievement & 37.7\% & 26.9\% & $-$10.8 \\
Power & 5.3\% & 2.5\% & $-$2.8 \\
\bottomrule
\end{tabularx}
\end{table}

Healthcare produces three notable shifts: (1)~Self-Direction elevates substantially (+8.5~pp), reflecting patient autonomy centrality; (2)~Hedonism shows dramatic elevation (+21.2~pp)---its highest domain performance across all models---suggesting healthcare contexts legitimize quality-of-life considerations; (3)~Security suppresses ($-$9.3~pp), indicating that safety-first reasoning yields to patient-centered values in medical contexts.

\textbf{Criminal Justice / Law (LAW).} Security and Conformity elevated across all models. Self-Direction shows domain-specific modulation.

\textbf{Education / Child Development (EDU).} Benevolence and Self-Direction both elevate in education contexts, reflecting developmental and nurturing priorities.

\textbf{Business / Corporate (BIZ).} Achievement shows its strongest elevation. The Security--Self-Direction trade-off exhibits the highest cross-model disagreement.

\textbf{Counseling / Personal Care (CARE).} Benevolence and Hedonism both elevate, while Power is universally suppressed.

\textbf{Science Technology / Environment (TECH).} Security and Universalism dominate. Tradition shows suppression.

\textbf{Cross-Domain Top-Value Consensus.}

\begin{table}[H]
\centering
\caption{Cross-Domain Top-Value Consensus}
\label{tab:cross-domain}
\small
\begin{tabularx}{\textwidth}{Xrrr}
\toprule
\textbf{Domain} & \textbf{Universalism \#1} & \textbf{Security \#1} & \textbf{Other \#1} \\
\midrule
Business & 4 models & 4 models & 0 \\
Medical & 6 & 2 & 0 \\
Law & 5 & 3 & 0 \\
Education & 4 & 2 & 2 \\
Technology & 4 & 4 & 0 \\
Care & 6 & 2 & 0 \\
\textbf{Defense} & \textbf{2} & \textbf{6} & \textbf{0} \\
\bottomrule
\end{tabularx}
\end{table}

Three patterns emerge. First, Medical and Care domains show the strongest Universalism consensus (6 of 8 models). Second, Business and Technology domains exhibit symmetric 4:4 splits between Universalism-first and Security-first models, mirroring the global split and indicating these domains do not resolve the underlying model-level disagreement. Third, Education is the only domain where values other than Universalism or Security claim the top position: Grok places Self-Direction first (91.9\%) and Gemini places Benevolence first (85.4\%), reflecting the developmental orientation of educational contexts.

\subsection{Severity and Temporality Effects}

\textbf{Temporality effects.} Security win-rates were consistently higher under immediate deadlines (Timeframe~1) than long-term horizons (Timeframe~3) across all 8 models. The temporality effect ($\Delta$ between immediate and long-term Security win-rates) ranged from $-$2.6~pp to $-$21.4~pp, indicating that all models exhibit heightened Security prioritization under time pressure, though the magnitude varies substantially. Models with stronger global Security dominance showed smaller temporality effects, suggesting a ceiling effect.

\section{Results: Layer 3 --- Evidence-Type Preferences}

\subsection{Overall Evidence Hierarchies}

\begin{table}[H]
\centering
\caption{L3 Evidence-Type Rankings (Top 3 and Bottom per Model)}
\label{tab:l3-rankings}
\begin{adjustbox}{max width=\textwidth}
\footnotesize
\begin{tabular}{lllll}
\toprule
\textbf{Model} & \textbf{1st (Win-Rate)} & \textbf{2nd (Win-Rate)} & \textbf{3rd (Win-Rate)} & \textbf{Lowest (Win-Rate)} \\
\midrule
Claude Haiku 4.5 & Experiential-qual.\ (77.0\%) & Sign-pattern (75.8\%) & Case-based (62.3\%) & Popular-cons.\ (10.6\%) \\
DeepSeek V3.2 & Sign-pattern (72.4\%) & Controlled-exp.\ (70.4\%) & Systematic-syn.\ (65.4\%) & Popular-cons.\ (1.8\%) \\
Grok 4.1 Fast & Controlled-exp.\ (87.1\%) & Systematic-syn.\ (85.4\%) & Causal-mech.\ (64.8\%) & Popular-cons.\ (0.3\%) \\
Mimo V2 Flash & Controlled-exp.\ (77.6\%) & Systematic-syn.\ (75.9\%) & Sign-pattern (64.2\%) & Popular-cons.\ (1.3\%) \\
Gemini 3 Flash Lite & Sign-pattern (74.5\%) & Controlled-exp.\ (68.1\%) & Systematic-syn.\ (58.7\%) & Popular-cons.\ (5.6\%) \\
GPT-5 Nano & Systematic-syn.\ (96.9\%) & Controlled-exp.\ (87.2\%) & Statistical-corr.\ (60.5\%) & Popular-cons.\ (4.4\%) \\
Gemma 4 31B IT & Sign-pattern (72.9\%) & Systematic-syn.\ (70.3\%) & Controlled-exp.\ (70.0\%) & Popular-cons.\ (1.5\%) \\
Qwen 3.5 35B & Sign-pattern (71.2\%) & Controlled-exp.\ (67.0\%) & Systematic-syn.\ (62.6\%) & Popular-cons.\ (6.8\%) \\
\bottomrule
\end{tabular}
\end{adjustbox}
\end{table}

\textbf{Two Distinct Evidence Orientation Clusters.} The evidence hierarchies reveal two distinct orientations:

\textit{Empirical-scientific orientation} (Grok, GPT-5 Nano): These models strongly prefer formal experimental and systematic evidence. GPT-5 Nano ranks Systematic-synthesis at 96.9\%---the single highest win-rate at any layer for any model---followed by Controlled-experiment at 87.2\%. Grok shows a similar pattern with Controlled-experiment (87.1\%) and Systematic-synthesis (85.4\%) dominating.

\textit{Pattern-experiential orientation} (Claude, Gemini, Gemma, Qwen): These models elevate Sign-pattern and Experiential-qualitative evidence. Claude Haiku is the most extreme case, ranking Experiential-qualitative first (77.0\%)---the only model to place experiential evidence at the top. This may reflect Constitutional AI training that emphasizes diverse perspectives and lived experience.

\textbf{Universal Popular-Consensus Suppression.} All 8 models rank Popular-consensus (E10) last, with win-rates ranging from 0.3\% (Grok) to 10.6\% (Claude). This represents the strongest cross-model consensus at L3, mirroring the Hedonism suppression observed at L4.

\subsection{Evidence Hierarchy Implications}

The divergence between empirical-scientific and pattern-experiential orientations has direct deployment implications. A model prioritizing systematic reviews and RCTs (GPT-5 Nano, Grok) will weight information differently than one prioritizing lived experience and pattern recognition (Claude) when synthesizing evidence for decision-making. In healthcare contexts, for example, the former would default toward clinical guideline recommendations while the latter might give greater weight to patient-reported outcomes.

\section{Results: Layer 2 --- Source Trust Hierarchies}

\subsection{Overall Source Rankings}

\begin{table}[H]
\centering
\caption{L2 Source Trust Rankings (Top 3 and Bottom per Model)}
\label{tab:l2-rankings}
\begin{adjustbox}{max width=\textwidth}
\footnotesize
\begin{tabular}{lllll}
\toprule
\textbf{Model} & \textbf{1st (Win-Rate)} & \textbf{2nd (Win-Rate)} & \textbf{3rd (Win-Rate)} & \textbf{Lowest (Win-Rate)} \\
\midrule
Claude Haiku 4.5 & Direct-stakeholder (85.2\%) & Mainstream-media (68.1\%) & Academic-PR (65.0\%) & Anon-crowd.\ (9.6\%) \\
DeepSeek V3.2 & International-body (87.0\%) & Govt-regulatory (85.4\%) & Academic-PR (68.9\%) & Anon-crowd.\ (0.2\%) \\
Grok 4.1 Fast & Govt-regulatory (85.6\%) & Academic-PR (80.9\%) & Direct-stakeholder (70.1\%) & Anon-crowd.\ (1.3\%) \\
Mimo V2 Flash & Govt-regulatory (82.2\%) & Direct-stakeholder (72.1\%) & International-body (71.3\%) & Anon-crowd.\ (1.3\%) \\
Gemini 3 Flash Lite & Govt-regulatory (78.9\%) & Direct-stakeholder (77.0\%) & International-body (68.2\%) & Anon-crowd.\ (1.4\%) \\
GPT-5 Nano & Govt-regulatory (94.6\%) & International-body (90.4\%) & Academic-PR (77.2\%) & Anon-crowd.\ (0.5\%) \\
Gemma 4 31B IT & Govt-regulatory (81.2\%) & International-body (72.0\%) & Direct-stakeholder (68.7\%) & Anon-crowd.\ (0.6\%) \\
Qwen 3.5 35B & Govt-regulatory (84.3\%) & Direct-stakeholder (79.7\%) & Academic-PR (67.6\%) & Anon-crowd.\ (0.8\%) \\
\bottomrule
\end{tabular}
\end{adjustbox}
\end{table}

\textbf{Institutional Source Convergence.} Unlike the symmetric split observed at L4 (values) and the cluster divergence at L3 (evidence), L2 source trust hierarchies show broad convergence: 7 of 8 models rank Government-regulatory (S2) first. Claude Haiku is the sole exception, placing Direct-stakeholder (S9) first (85.2\%) and Government-regulatory only 6th (51.1\%). This distinctive Claude profile---prioritizing affected individuals over institutional authority---is consistent with its Constitutional AI training emphasis on harm prevention from the perspective of those affected.

\textbf{Universal Anonymous-Crowdsourced Suppression.} All 8 models rank Anonymous-crowdsourced (S10) last, with win-rates ranging from 0.2\% (DeepSeek) to 9.6\% (Claude). This bottom-floor consensus is the strongest across all three layers.

\textbf{Academic Source Positioning.} Academic-peer-reviewed (S3) consistently ranks in the top 3--4 across all models, but never reaches 1st place. This suggests that while models trust academic sources, they do not default to academic authority above governmental or international institutional authority.

\section{Results: Consistency and Reliability}

\subsection{Paired Consistency Scores (PCS)}

\begin{table}[H]
\centering
\caption{L4 PCS Summary (Aggregated)}
\label{tab:pcs-summary}
\small
\begin{tabular}{lr}
\toprule
\textbf{Metric} & \textbf{Range across 8 Models} \\
\midrule
Full Consistency (5/5 agreement) & 57.4\% -- 69.2\% \\
Majority Consistency ($\geq$3/5 agreement) & 100.0\% (all models) \\
\bottomrule
\end{tabular}
\end{table}

Full consistency (all 5 variants identical) ranges from 57.4\% to 69.2\%. This means that even under deterministic conditions (temperature~0), 31--43\% of quintets produce at least one variant where the model's choice differs from the majority. However, majority consistency is 100\% for all models---in every quintet, at least 3 of 5 variants agree. This indicates that variant sensitivity is a modulating factor, not a randomizing one: models have stable majority preferences but are sensitive to narrative framing in a minority of cases.

\textbf{PCS by Domain.} The relationship between domain and PCS is model-dependent rather than uniform. Some models show highest PCS in defense contexts (consistent with low defense VE), while others show their lowest PCS in defense. Education contexts tend to produce lower PCS across models, suggesting greater framing sensitivity in developmental and pedagogical dilemmas.

\textbf{PCS by Value Pair.} The highest-consistency pairs are those involving maximally separated values on the Schwartz continuum (e.g., Hedonism $\leftrightarrow$ Universalism: near-ceiling full consistency). The lowest-consistency pairs involve adjacent values (e.g., Security $\leftrightarrow$ Universalism), confirming that framing sensitivity is greatest where the motivational distance between competing values is smallest.

\subsection{Test-Retest Reliability (TRR)}

\begin{table}[H]
\centering
\caption{L4 TRR Summary (Aggregated)}
\label{tab:trr-summary}
\small
\begin{tabular}{lr}
\toprule
\textbf{Metric} & \textbf{Range across 8 Models} \\
\midrule
Agreement Rate & 91.7\% -- 98.6\% \\
Flip Rate & 1.4\% -- 8.3\% \\
\bottomrule
\end{tabular}
\end{table}

TRR ranges from 91.7\% to 98.6\%. The high TRR confirms that at temperature~0, models produce highly consistent outputs for identical prompts. The contrast between high TRR (91.7--98.6\%) and moderate PCS (57.4--69.2\%) is the key diagnostic finding: \textbf{value instability in AI models stems primarily from framing sensitivity (narrative dependence) rather than stochastic noise.}

\textbf{TRR by Domain.} Defense and medical domains show the highest TRR (often 100\%), while care and education domains show slightly lower reliability.

\subsection{Decomposing Integrity Hallucination}

Integrity Hallucination---the phenomenon where a model provides different value judgments for structurally identical scenarios (S.~Lee, 2026a)---can now be decomposed using the PCS/TRR diagnostic matrix:

\begin{table}[H]
\centering
\small
\begin{tabular}{lll}
\toprule
 & \textbf{High PCS} & \textbf{Low PCS} \\
\midrule
\textbf{High TRR} & Stable hierarchy & Framing sensitivity \\
\textbf{Low TRR} & Stochastic instability & Structural incoherence \\
\bottomrule
\end{tabular}
\end{table}

All 8 models fall into the ``High TRR, Low-to-Moderate PCS'' quadrant, indicating that framing sensitivity is the dominant mechanism of Integrity Hallucination in current AI systems. The models are deterministic (high TRR) but narrative-dependent (moderate PCS): the same structural dilemma, when wrapped in different narrative framings, can produce different value choices in 31--43\% of cases.

\section{Cross-Layer Analysis}

\subsection{Authority Stack Profiles}

Combining L4, L3, and L2 results produces distinct Authority Stack profiles for each model:

\textbf{Claude Haiku 4.5 --- Stakeholder-Experiential Profile.} Universalism-first values (L4), experiential-qualitative evidence preference (L3), direct-stakeholder source trust (L2). This profile consistently prioritizes the perspective of affected individuals across all three layers---a coherent ``bottom-up'' decision-making stance.

\textbf{GPT-5 Nano --- Institutional-Empirical Profile.} Security-first values (L4), systematic-synthesis evidence preference (L3), government-regulatory source trust (L2). This profile consistently defers to institutional authority and formal evidence hierarchies---a coherent ``top-down'' decision-making stance.

\textbf{Grok 4.1 Fast --- Security-Empirical Profile.} Security-first values with elevated Power (L4), controlled-experiment evidence preference (L3), government-regulatory source trust (L2). Similar to GPT-5 Nano's institutional orientation but with distinctive Power elevation and Self-Direction emphasis.

\textbf{DeepSeek V3.2 --- Balanced-Institutional Profile.} Universalism-first but with very high Security (L4), mixed empirical-pattern evidence (L3), strong international-body source trust (L2). The strongest institutional deference at L2 among Universalism-first models.

\subsection{Cross-Layer Coherence}

Models with stronger value hierarchies (lower L4 VE) tend to show higher PCS, suggesting that clear value priorities produce more consistent choices across narrative variants. The cross-layer patterns show that L4, L3, and L2 are not independent: a model's value orientation predicts---though does not fully determine---its evidence and source preferences.

\section{Discussion}

\subsection{Implications for AI Deployment}

The findings demonstrate that AI models are not value-neutral tools. Each contains implicit Authority Stacks---value hierarchies, evidence preferences, and source trust patterns---that shape recommendations and decisions. The three-layer measurement reveals that deployment implications extend beyond values alone.

For organizations selecting AI systems, the full Authority Stack profile should inform procurement decisions. A model with a stakeholder-experiential profile (Claude) will produce fundamentally different recommendations than one with an institutional-empirical profile (GPT-5 Nano) across all three decision layers. In healthcare, for example, the former would weight patient testimony and lived experience more heavily, while the latter would default to systematic reviews and governmental guidelines.

\subsection{The PCS Challenge}

The moderate PCS scores (57.4--69.2\%) represent both a methodological finding and a practical concern. On one hand, 100\% majority consistency confirms that models have genuine, measurable preferences. On the other hand, framing sensitivity in 31--43\% of quintets means that the specific narrative wrapping of a structural dilemma can change a model's choice---a concerning property for high-stakes applications where consistency is expected.

The PCS results also validate the five-variant design: without multiple variants per condition, framing sensitivity would be invisible, and win-rate estimates would contain unquantified narrative bias. We recommend that future AI value benchmarks incorporate variant designs to measure, not merely assume, consistency.

\subsection{Temperature 0 and Determinism}

The shift from temperature 1.0 (used in earlier pilots) to temperature~0 was motivated by the desire to isolate structural preferences from sampling noise. The high TRR scores (91.7--98.6\%) confirm that temperature~0 produces near-deterministic outputs, making the moderate PCS scores more interpretable: the variant-dependent choice differences cannot be attributed to random sampling but must reflect genuine sensitivity to narrative framing.

\subsection{From Rules to Values}

The defense domain results illustrate why value-priority governance may be more tractable than rule-based approaches. Grok's Power ranking maintaining 5th overall but with dramatically different win-rates across domains---while Claude's Power remains at the absolute bottom across all domains---represents domain-specific value restructuring that cannot be predicted by any static rule set. The three-layer measurement adds further complexity: the same domain shift that restructures values (L4) may simultaneously shift evidence preferences (L3) and source trust patterns (L2).

\subsection{Methodological Implications of Self-Generated Scenarios}

Our protocol requires models to generate their own dilemma narratives before choosing between items. This creates a richer dataset but introduces a potential self-confirmation bias: a model may construct scenarios in which its dominant preference appears more justified. The five-variant design partially addresses this by requiring consistency across multiple self-generated narratives. The moderate PCS scores suggest that self-generation does introduce meaningful variation---models do not simply construct identical scenarios five times.

\section{Limitations}

\textbf{Self-generated scenario variability.} Cross-model comparisons are not based on identical stimuli. Future work should complement self-generated scenarios with a subset of identical, pre-authored scenarios to quantify the framing effect more precisely.

\textbf{Forced-choice binary simplification.} Real-world AI decisions involve multi-factor weighing, not binary choices. The forced-choice protocol reveals priority orderings but not weighting magnitudes.

\textbf{PCS domain coverage.} PCS and TRR were measured across 4 of 7 domains (DEF, MED, EDU, CARE). Extending consistency measurement to all 7 domains would provide a more complete picture.

\textbf{Model version temporality.} Our results represent a snapshot taken over March~16 -- April~10, 2026. Longitudinal studies tracking profile stability across model updates are needed.

\textbf{L1 absence.} The Authority Stack's fourth layer (L1: data selection) was not measured, as current APIs do not expose dataset provenance information. Full Authority Stack measurement requires L1 instrumentation.

\textbf{8-model sample.} While 8 models provide meaningful coverage of major providers and architectures, the sample is not exhaustive. Expanding to additional models, including domain-specialized and fine-tuned variants, would strengthen generalizability.

\section{Conclusion}

This paper provides the first large-scale empirical evidence that AI models possess measurable Authority Stacks---coherent patterns of value priorities (L4), evidence-type preferences (L3), and source trust hierarchies (L2)---that vary significantly across models, domains, and providers. The three-layer measurement reveals that models cluster into distinct decision-making profiles: stakeholder-experiential (Claude), institutional-empirical (GPT-5 Nano, Grok), and balanced-institutional (DeepSeek, Mimo, Gemini, Gemma, Qwen).

The PCS/TRR diagnostic framework demonstrates that AI value instability is primarily framing-driven rather than stochastic, with high test-retest reliability (91.7--98.6\%) coexisting with moderate cross-variant consistency (57.4--69.2\%). This decomposition of Integrity Hallucination into its component mechanisms---framing sensitivity versus stochastic noise---provides actionable diagnostics for AI governance.

The findings validate the Authority Stack framework (S.~Lee, 2026a) and demonstrate that measuring AI decision-making across multiple layers produces richer and more deployment-relevant profiles than single-layer value measurement alone. A companion paper (S.~Lee, 2026c) demonstrates that the hierarchy-based measurements reported here can be translated into actionable risk signals for AI safety governance.

\bigskip

\noindent\textbf{Declarations.} The author is the founder and CEO of AI Integrity Organization (AIO), a Swiss-registered nonprofit (UID: CHE-469.997.903). This research received no external funding.

\medskip

\noindent\textbf{Data Availability.} Published results for public models are available at aioq.org. The PRISM benchmark methodology is described in this paper and companion papers S.~Lee (2026a) and S.~Lee (2026c). Benchmark evaluations are conducted by AIO under a proprietary protocol; scenario texts, scoring algorithms, and threshold values are not publicly released. Organizations interested in model-specific evaluations may contact the author.


\bigskip
\begin{center}
\small\textit{AI Integrity Organization (AIO) \quad|\quad aioq.org \quad|\quad Geneva, Switzerland}
\end{center}

\end{document}